\documentclass[letterpaper, 10 pt, conference]{ieeeconf}  

\IEEEoverridecommandlockouts                              

\usepackage{graphics} 
\usepackage{epsfig} 
\usepackage{mathptmx} 
\usepackage{times} 
\usepackage{amsmath} 
\usepackage{amssymb}  
\usepackage{xcolor}
\usepackage{colortbl}
\usepackage{algorithm}
\usepackage{amsfonts}
\usepackage{relsize}
\usepackage{booktabs}
\usepackage{multirow}
\usepackage{pifont}
\usepackage{booktabs}
\usepackage{pifont}
\usepackage{makecell}
\usepackage{soul}
\usepackage[pagebackref=false,colorlinks]{hyperref}
\soulregister{\ref}7
\soulregister{\cite}7

\newcommand{\cmark}{\ding{51}}%

\newcommand{\method}{IR-WM}%

\usepackage{newfloat}
\usepackage{listings}

\title{\LARGE \bf
Vision-Centric 4D Occupancy Forecasting and Planning via \\ Implicit Residual World Models
}

\author{Jianbiao Mei$^{*}$, Yu Yang$^{*}$, Xuemeng Yang$^{\dagger}$, Licheng Wen, Jiajun Lv$^{\dagger}$, Botian Shi, Yong Liu
\thanks{* Equal contributors. $\dagger$ Corresponding authors.}
\thanks{
Jianbiao Mei, Yu Yang, Jiajun Lv, and Yong Liu are with Zhejiang University.
Xuemeng Yang, Licheng Wen, and Botian Shi are with Shanghai AI Laboratory. }
}
\begin{document}

\maketitle
\thispagestyle{empty}
\pagestyle{empty}

\begin{abstract}
End-to-end autonomous driving systems increasingly rely on vision-centric world models to understand and predict their environment. However, a common ineffectiveness in these models is the full reconstruction of future scenes, which expends significant capacity on redundantly modeling static backgrounds. To address this, we propose {\method}, an Implicit Residual World Model that focuses on modeling the current state and evolution of the world.
{\method} first establishes a robust bird's-eye-view representation of the current state from the visual observation. It then leverages the BEV features from the previous timestep as a strong temporal prior and predicts only the ``residual", i.e., the changes conditioned on the ego-vehicle’s actions and scene context. To alleviate error accumulation over time, we further apply an alignment module to calibrate semantic and dynamic misalignments. Moreover, we investigate different forecasting–planning coupling schemes and demonstrate that the implicit future state generated by world models substantially improves planning accuracy. On the nuScenes benchmark, {\method} achieves top performance in both 4D occupancy forecasting and trajectory planning. Codes are available at \url{https://github.com/yuyang-cloud/Drive-OccWorld}
\end{abstract}

\section{Introduction}
End-to-end autonomous driving models, which directly map raw sensor data to planning outputs, have become a dominant paradigm~\cite{hu2023planning, jiang2023vad}. At its core, safe decision-making in autonomous driving hinges on the precise understanding of current and past status, along with the reliable anticipation of future events. Consequently, action-driven world models have emerged as a promising approach that bridges agent intentions and environmental dynamics through action-guided world generation and forecast-driven planning, thereby enabling more reliable decision-making. 

\begin{figure}[t]
\centering
	\includegraphics[width=\linewidth]{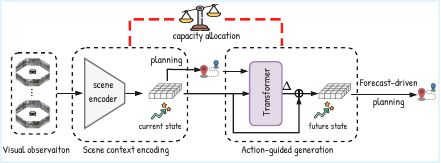}
    \vspace{-20pt}
	\caption{ We model the current state and predict residual future changes, yielding a more effective allocation of model capacity.}
	\label{fig:teaser}
 \vspace{-1.5em}
\end{figure}

However, existing vision-centric world models for end-to-end planning often cast forecasting as full-scene reconstruction. Whether they are implicit models that produce low-dimension latents \cite{li2024navigation, li2024enhancing} or explicit models that predict high-dimension representations like videos and 4D occupancy \cite{min2024driveworld, yang2025driving}, they tend to regenerate the entire scene from scratch for both current and future states, wasting capacity on largely static backgrounds and constraining scene context encoding from past and current visual observation. This motivates a shift from regenerating the world to modeling the current state and its future changes, as shown in Fig. \ref{fig:teaser}, better allocating model capacity.

Therefore, we introduce the Implicit Residual World Model, \textbf{{\method}}, guided by auxiliary vision-centric 4D occupancy forecasting to enable action-guided world generation and forecast-driven planning. {\method} first constructs an implicit bird’s-eye-view (BEV) representation of the current state from raw images. Instead of reconstructing full futures, it treats the previous BEV as a spatiotemporal prior and predicts only residual changes conditioned on ego trajectories. This residual design yields: (1) modeling efficiency by focusing capacity on scene context encoding from visual observation and salient context changes, e.g., dynamics and background changes, for forecasting; (2) improved dynamics understanding by emphasizing temporal differences; and (3) temporal coherence by grounding each prediction in the preceding state. We recover future states by adding residuals to the prior and applying an alignment module to correct semantic and dynamic misalignments, alleviating potential error accumulation during rollout. All BEV representations are supervised with semantic occupancy grids, enhancing semantic richness and geometric fidelity for reliable planning.

Moreover, we design different forecasting–planning coupling schemes to study how forecasting influences planning. Our insights are: (1) using occupancy forecasting to filter candidate trajectories is largely unnecessary, offering only marginal planning gains while adding latency; (2) implicit future state generated by world models substantially improves planning accuracy; and (3) action-conditioned world models can serve as effective observers, providing strong supervisory signals for trajectory generation.


We evaluate {\method} on nuScenes~\cite{caesar2020nuscenes} for vision-centric 4D occupancy and trajectory planning. Under the OpenOccupancy benchmark~\cite{wang2023openoccupancy}, {\method} achieves consistent gains across 4D occupancy tasks and significantly reduces L2 error and collision rate relative to the baseline.

\begin{figure*}[t]
\centering
	\includegraphics[width=0.98\textwidth]{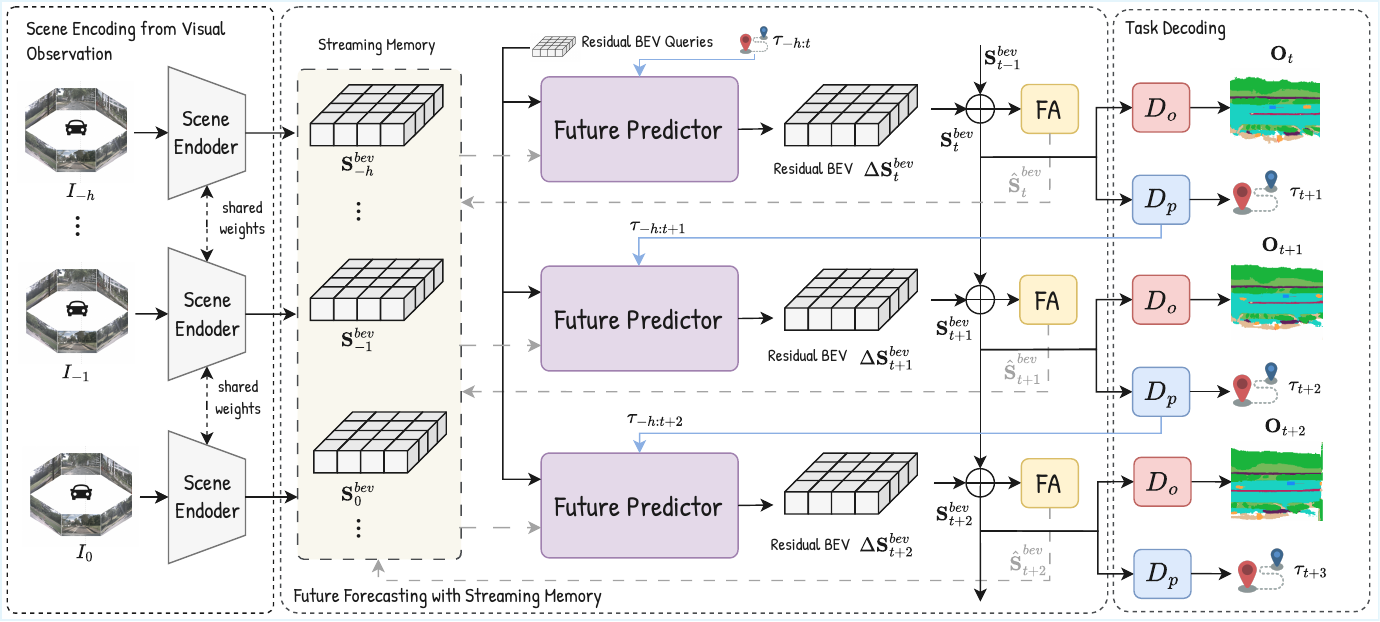}
    \vspace{-5pt}
	\caption{Overview of \method. {\method} comprises three parts: (a) Scene encoding from visual observations via a scene encoder; (b) future forecasting, performed in an implicit residual prediction manner with a streaming memory using an autoregressive future predictor; (c) Task decoding via occupancy head $D_o$ and planning head $D_p$. ``FA" denotes the feature alignment for alleviating error accumulation.}
	\label{fig:pipeline}
 \vspace{-1em}
\end{figure*}

Our main contributions can be summarized as follows:
\begin{itemize}
\item We introduce {\method}, a world model that models the current world state and its dynamics in a compact BEV space, providing a strong basis for vision-centric 4D occupancy forecasting and trajectory planning.
    \item We investigate the impact of forecasting on planning, and exploit a semi-coupled design separating 4D occupancy and planning to enhance flexibility while maintaining planning accuracy.
    \item Extensive experiments demonstrate the effectiveness of {\method}. It achieves state-of-the-art performance in both 4D occupancy forecasting and trajectory planning.

\end{itemize}

\section{Related Works}
\subsection{Driving World Models}
World models aim to simulate future environment states given past observations and actions. Image-based approaches \cite{hu2023gaia, wang2023drivedreamer,li2023drivingdiffusion, yang2024generalized, kong20253d, mei2024dreamforge, yang2025drivearena} generate future driving videos using diffusion or transformer architectures, often conditioned on actions, BEV layouts, or HD maps to enhance realism. In contrast, volume-based methods \cite{zhang2023learning, zheng2023occworld, wang2024occsora, min2024driveworld, xu2025t3former, yang2025x} directly model 3D structures such as occupancy or point clouds, enabling spatially consistent future prediction. Unlike recent approaches, we construct a residual BEV world model with feature alignment, guided by 4D occupancy forecasting, which balances model capacity to effectively capture both the current state and its future changes. This design alleviates error accumulation, leverages vision-centric state modeling as a foundation for accurate future prediction, and strengthens downstream planning.

\subsection{End-to-end Driving}
End-to-end models have emerged as a paradigm shift in autonomous driving, directly mapping sensor inputs to future trajectories \cite{hu2023planning, jiang2023vad}. Early approaches such as P3 \cite{sadat2020perceive} and ST-P3 \cite{hu2022st} incorporated differentiable occupancy to ensure safe planning. More recent frameworks leverage BEV perception as an intermediate representation, including UniAD \cite{hu2023planning}, VAD \cite{jiang2023vad}, and GraphAD \cite{zhang2024graphad}. Further advances focus on disentangled scene representations \cite{doll2024dualad}, reducing annotation costs \cite{guo2024end, li2024navigation}, or improving efficiency through parallel or sparse architectures \cite{weng2024drive, zhang2024sparsead}. Our work explores the impact of implicit representation generated by world models on end-to-end planning.

\subsection{Occupancy Prediction and Forecasting}
Occupancy prediction reconstructs the 3D state of the environment. LiDAR-based methods \cite{song2017semantic, yan2021sparse, yang2021semantic, mei2023ssc} complete sparse point clouds into dense voxel grids, while camera-based approaches \cite{li2023voxformer, wei2023surroundocc, wang2023openoccupancy, mei2024camera} transform 2D features into 3D space, often enhanced with depth estimation \cite{zhang2023occformer} or multi-view aggregation \cite{tian2023occ3d}. Forecasting extends this task to the temporal domain, with LiDAR-driven approaches that predict future points or occupancy volumes \cite{khurana2022differentiable, luo2023pcpnet}, and 4D occupancy forecasting of Cam4DOcc \cite{ma2024cam4docc} and Drive-OccWorld \cite{yang2025driving}. This work leverages occupancy forecasting as a fine-grained regularization to learn compact world state representations and to generate state-conditioned trajectories.

\section{Method}
\subsection{Preliminary}
Formally, letting $t=0$ denote the current timestamp, the vision-centric end-to-end planner takes the past and current camera observations $\mathbf{I}_{-h:0}$ and ego trajectories $\mathbf{\tau}_{-h:0} \in \mathbb{R}^{(h+1) \times 2}$ over past $h$ and current timestamps, and predicts the ego trajectories $\mathbf{\tau}_{1:f} \in \mathbb{R}^{f \times 2}$ over the next $f$ steps:
\begin{equation}
    \mathbf{\tau}_{1:f} = \mathcal{F}(I_{-h:0}, \tau_{-h:0})
\end{equation}
    
End-to-end planning with world models typically decomposes this procedure into three stages: compact scene context encoding from camera observations, action-conditioned future state rollout, and downstream perception and planning:
\begin{equation}
     \mathbf{S}_{-h:0} = \mathcal{E}(\mathbf{I}_{-h:0}, \mathbf{\tau}_{-h:0})
\end{equation}
\begin{equation}
    \mathbf{S}_{t} = \mathcal{W}(\mathbf{S}_{-h:t-1}, \mathbf{\tau}_{-h:t})
\end{equation}
\begin{equation}
    \mathbf{\tau}_{t+1} = \mathcal{D}_p(\mathbf{S}_{t})
\end{equation} where $\mathbf{S}$ denotes the compact state representations (e.g., BEV features). $\mathcal{E}$ is the scene encoder, $\mathcal{W}$ is the autoregressive future predictor, and $\mathcal{D}_{*}$ is the task decoder, such as the planning head. In this work, we also append an occupancy head to the model that (i) provides fine-grained perception outputs for interpretability and (ii) regularizes the latent state representations, thereby enhancing the semantic richness and geometric fidelity of both current and future states for reliable planning. The occupancy prediction for timestamp $t$ is formulated as:
\begin{equation}
    \mathbf{O}_{t} = \mathcal{D}_o(\mathbf{S}_{t})
\end{equation}


Accurate rollout hinges on informative encodings of both history and the current scene. Yet many methods rebuild the entire scene at each step, overloading $\mathcal{W}$ with static background prediction and limiting $\mathcal{E}$ ’s capacity for contextual encoding. We therefore propose {\method}, which explicitly models the current world state and its residual evolution, providing a strong basis for 4D occupancy forecasting and trajectory planning.

\subsection{Implicit Residual World Model}
As depicted in Figure \ref{fig:pipeline}, {\method} comprises three parts: \textbf{(1)} Scene encoding from visual observations via a scene encoder $\mathcal{E}$. \textbf{(2)} Future forecasting with streaming memory via autoregressive future predictor $\mathcal{W}$. \textbf{(3)} Task decoding via occupancy head $\mathcal{D}_o$ and planning head $\mathcal{D}_p$.

\subsubsection{Scene Encoding from Visual Observation} Following previous works \cite{yang2024visual, min2024driveworld}, we utilize the BEVFormer \cite{li2022bevformer} as our scene encoder $\mathcal{E}$ to take historical and current camera images as input, extract multi-view scene context, and transform them into BEV features $\mathbf{S}^{bev}_{-h:0}$, which are push into the streaming memory for future forecasting.

\subsubsection{Future Forecasting with Streaming Memory} 

We build an autoregressive predictor $\mathcal{W}$ that models context changes from streaming memory and ego trajectories, and use an alignment module to correct semantic and dynamic misalignments in future BEV features. $\mathcal{W}$ consists of a stack of transformer layers and takes learnable BEV queries as input. At timestamp $t$, each layer performs deformable self-attention, deformable temporal cross-attention to memory $\mathbf{S}^{bev}_{t-m:t-1}$ (with $m$ memory frames), conditioning via ego-trajectory embeddings $\tau_{t-m:t}$, and an FFN to output the residual $\Delta \mathbf{S}^{bev}_{t}$ w.r.t. the previous BEV. For efficiency, temporal cross-attention uses deformable sampling; reference points are computed from ego-motion transforms $\delta\tau_{t}$ derived from $\tau_{t-1:t}$ to pre-align features across time before aggregation.

\textbf{Feature Alignment.} To obtain the final BEV features $\mathbf{S}^{bev}_{t}$, we add the predicted feature differences $\Delta\mathbf{S}^{bev}_{t}$ with previous BEV features $\mathbf{S}^{bev}_{t-1}$. Then we apply a feature alignment module further to enhance both the semantic richness and dynamic understanding, providing aligned features for the next rollout to alleviate potential error accumulation.
Inspired by AdaNorm \cite{xu2019understanding} and \cite{yang2025driving}, instead of just adding the information, we modulate the features by dynamically generating the scale and shift parameters ($\gamma, \beta$) of layer normalization. 
The scale and shift parameters $(\gamma_s, \beta_s)$ for semantics and $(\gamma_e, \beta_e)$ for dynamics are generated by separate MLPs from the predicted semantic occupancy maps $\mathbf{O}^{align}_{t}$ and ego-motion transformations $\delta\tau_{t}$, respectively.
The procedure can be formulated as:
\begin{equation}
    \mathbf{S}^{bev}_{t} = \Delta\mathbf{S}^{bev}_{t} + \mathbf{S}^{bev}_{t-1}
\end{equation}
\begin{align}
    \mathbf{\hat{S}}^{bev}_{t} = \gamma_s(\mathbf{O}^{align}_{t})\cdot\mathrm{LN}(\mathbf{S}^{bev}_{t}) + \beta_s(\mathbf{O}^{align}_{t}) + \\
    \gamma_e(\delta\tau_{t})\cdot\mathrm{LN}(\mathbf{S}^{bev}_{t}) + \beta_e(\delta\tau_{t})
\end{align} where $\rm{LN}$ denotes the layer normalization.

We append a lightweight head consisting of MLPs to BEV features $\mathbf{S}^{bev}_{t}$ to generate a semantic occupancy map $\mathbf{O}^{align}_{t}$, which is supervised by cross-entropy loss $\mathcal{L}_{align}$. After obtaining aligned BEV features $\mathbf{\hat{S}}^{bev}_{t}$, we push them into the streaming memory queue for the next rollout. 

\begin{figure}[t]
\centering
	\includegraphics[width=0.85\linewidth]{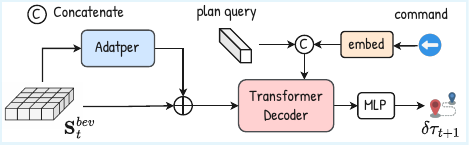}
	\caption{The detailed architecture of the planning head.}
	\label{fig:sp}
 \vspace{-1em}
\end{figure}

\subsubsection{4D Occupancy Forecasting and Planning}
After obtaining the BEV features $\mathbf{S}^{bev}_{t}$, we employ two parallel task-specific heads to support both perception and decision making: an \emph{occupancy head} for 4D occupancy forecasting and a \emph{planning head} for trajectory generation.

\textbf{Occupancy Head.}
Following \cite{yu2023flashocc, mei2023ssc}, an MLP-based occupancy head applies a channel-to-height mapping to lift BEV features into a vertical volume, producing dense semantic occupancy $\mathbf{O}_{t}$, where each voxel encodes geometry and class. This explicit representation enhances interpretability and geometric fidelity. Rolling out over multiple steps yields $\mathbf{O}_{0:f-1}$ for current and future frames, capturing the scene’s 4D spatiotemporal evolution.

\textbf{Planning Head.}
As shown in Fig. \ref{fig:sp}, for planning, a learnable plan query predicts the ego translation $\delta\tau_{t+1}$, updating the pose as $\tau_{t+1} = \tau_t + \delta\tau_{t+1}$.  At timestamp $t$, the plan query is concatenated with the command embedding and passed to a transformer decoder that performs cross-attention over $\textbf{S}_t^{bev}+\rm{Adapter}(\textbf{S}_t^{bev})$ to capture fine-grained scene context and attend to potential dynamic obstacles, followed by a feed-forward network; an MLP then outputs $\delta\tau_{t+1}$. The adapter is a stack of convolutions for feature enhancement.

\subsubsection{Interaction between Forecasting and Planning} 
\begin{figure}[t]
\centering
	\includegraphics[width=0.95\linewidth]{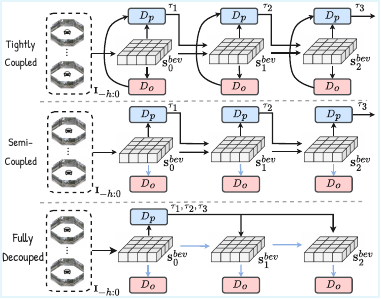}
	\caption{Three variants illustrating how forecasting affects planning; the blue line marks components that can be removed during inference.}
	\label{fig:planner}
 \vspace{-1em}
\end{figure}

Moreover, we design several variants to examine how forecasting affects planning, as shown in Fig. \ref{fig:planner}:

\textbf{Tightly Coupled}: Planning relies not only on generated BEV features $\mathbf{S}^{bev}_{0:f-1}$ but also on predicted semantic occupancy $\mathbf{O}_{0:f-1}$. Following ST-P3 \cite{hu2022st} and Drive-OccWorld \cite{yang2025driving}, at timestamp $t$, the predicted semantic occupancy $\mathbf{O}_{t}$ is used to filter initial candidate trajectories $\mathbf{\tau}_{t+1}^*$, which are guided by high-level commands and optimized via a safety-oriented cost function. The planner then predicts the final trajectory $\tau_{t+1}$ conditioned on the filtered candidates and the BEV features $\mathbf{S}^{bev}_{t}$, which can be formulated as:
\begin{equation}
    \tau^*_{t+1} = \rm{argmin}_{\tau_{sample}}\mathcal{C}(\tau_{sample}, \textbf{O}_{t}, c)
\end{equation}
\begin{equation}
    \tau_{t+1} = \mathcal{D}_p(\tau_{t+1}^*, \mathbf{S}^{bev}_{t}, c)
\end{equation} where $\mathcal{C}$ is the cost function in ST-P3 and $c$ denotes the high-level command. $\tau_{sample}$ is the sampled initial trajectories.

\textbf{Semi-Coupled}: Planning only depends on the generated BEV features $\mathbf{S}^{bev}_{0:f-1}$. Occupancy head is used only as auxiliary supervision and regularization, enhancing the learned BEV feature and providing optional interpretable perception results. Our {\method} adopts this strategy by default as suggested by experiments in Table \ref{ab:planner}.

\textbf{Fully Decoupled}: Trajectory prediction is performed solely on the current BEV features $\mathbf{S}^{bev}_{0}$, with future BEV features and semantic occupancy serving only as auxiliary supervision. Under this setting, no explicit forecasting of future features is required during inference:
\begin{equation}
    \tau_{1:f} = \mathcal{D}_p(\mathbf{S}^{bev}_{0}, c)
\end{equation}

\subsection{Training Objective}
We jointly optimize occupancy forecasting and trajectory planning with a combination of perception-oriented and planning-oriented losses.

\textbf{Semantic Occupancy Supervision.}
For occupancy prediction, we combine complementary objectives to balance semantic accuracy and geometric consistency:
\begin{equation}
    \mathcal{L}_{occ} = \frac{1}{f}\sum_{t=0}^{f-1} \big[
    \mathcal{L}_{ce}(\mathbf{O}_t, \hat{\mathbf{O}}_t) +
    \mathcal{L}_{lovasz}(\mathbf{O}_t, \hat{\mathbf{O}}_t) +
    \mathcal{L}_{bce}(\mathbf{O}_t, \hat{\mathbf{O}}_t)
    \big],
\end{equation}
where $f$ is the number of future steps, $\hat{\mathbf{O}}_t$ denotes the ground-truth semantic occupancy. The cross-entropy loss supervises voxel-wise semantic classification, the Lovász loss \cite{berman2018lovasz} directly optimizes IoU to improve segmentation quality, and binary occupancy loss emphasizes free/occupied space to preserve geometric fidelity.

\textbf{Planning Supervision.}
For trajectory planning, we employ a hybrid objective that integrates imitation learning and collision avoidance:
\begin{equation}
    \mathcal{L}_{plan} = \frac{1}{f}\sum_{t=1}^{f} \Big[
    |\tau_t - \hat{\tau}_t|_2^2 +
    \lambda_{coll} \cdot \mathcal{L}_{coll}(\hat{\tau}_t, \hat{\mathbf{O}}_t)
    \Big],
\end{equation}
where $|\cdot|_2^2$ is the trajectory regression loss, and $\mathcal{L}_{coll}$ \cite{hu2022st} penalizes predicted waypoints that overlap with occupied voxels in $\hat{\mathbf{O}}_t$. The hyperparameter $\lambda_{coll}$ balances fidelity of imitation learning with safety constraints.

\textbf{Joint Optimization.}
The final training objective is a weighted sum of perception and planning losses:
\begin{equation}
    \mathcal{L} = \mathcal{L}_{align} + \mathcal{L}_{occ} + \lambda_{plan} \cdot \mathcal{L}_{plan},
\end{equation}
where $\lambda_{plan}$ controls the trade-off between occupancy forecasting and trajectory planning. 

\begin{figure*}[t]
\centering
	\includegraphics[width=0.96\linewidth]{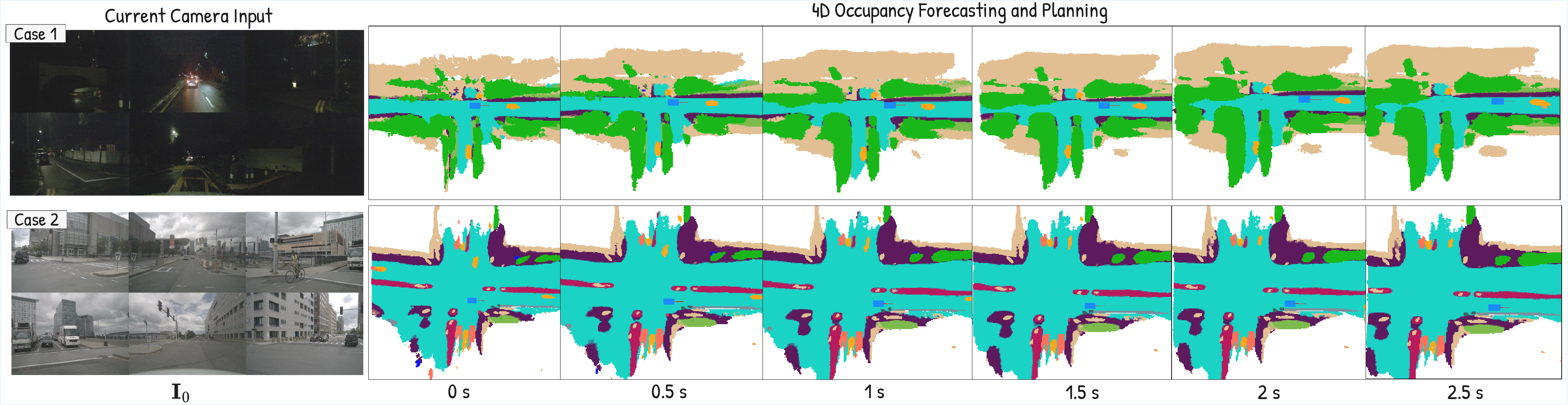}
    \vspace{-7pt}
	\caption{\textbf{Qualitative results for 4D occupancy forecasting and planning} show that our method yields reliable predictions for static and dynamic objects at different times of day.}
	\label{fig:visualization}
 \vspace{-1em}
\end{figure*}

\begin{table}[t]
    \centering
    \renewcommand\arraystretch{1}
    \definecolor{mycolor}{rgb}{0.64, 0.87, 0.93}
    \caption{Comparisons of \textbf{Inflated GMO Forecasting} on the nuScenes dataset, and \textbf{Fine-Grained GMO Forecasting} on the nuScenes-Occupancy dataset.}
    \renewcommand\tabcolsep{2pt}
    \scalebox{0.93}{
        \begin{tabular}{l|ccc|ccc}
        \toprule
         & \multicolumn{3}{c|}{nuScenes} & \multicolumn{3}{c}{nuScene-Occupancy} \\
         \cline{2-7}
         \multirow{-2}{*}{Method} & $\text{IoU}_{c}$ & $\text{IoU}_{f}$ (2\ s) & $\tilde{\text{IoU}}_{f}$ & $\text{IoU}_{c}$ & $\text{IoU}_{f}$ (2\ s) & $\tilde{\text{IoU}}_{f}$ \\
        \midrule
        OpenOccupancy-C \cite{wang2023openoccupancy} & 12.17 & 11.45 & 11.74 & 10.82 & 8.02 & 8.53 \\
        $\rm{SPC}^\dagger$ & 1.27 & - & - & 5.85 & 1.08 & 1.12 \\
        PowerBEV-3D \cite{li2023powerbev} & 23.08 & 21.25 & 21.86 & 5.91 & 5.25 & 5.49 \\
        BEVDet4D \cite{huang2022bevdet4d}& 31.60 & 24.87 & 26.87 & -& -& -\\
        OCFNet (Cam4DOcc) \cite{ma2024cam4docc} & 31.30 & 26.82 & 27.98 & 11.45 & 9.68 & 10.10 \\
        OccProphet \cite{chen2025occprophet} & 34.36 & 26.94 & 29.15 & 15.38 & 10.69 & 11.98 \\
        Drive-OccWorld \cite{yang2025driving} & 39.80 & 36.30 & 37.40 & 13.60 & 12.00 & 12.40 \\
        \hline
        \rowcolor{mycolor!25}
        {\method} (ours) & \textbf{40.80} & \textbf{37.20} & \textbf{38.20} & \textbf{16.20} & \textbf{14.50} & \textbf{15.00 }\\
        \bottomrule
        \multicolumn{7}{l}{$\rm{SPC}^\dagger$: SurroundDepth \cite{wei2023surrounddepth} + PCPNet \cite{luo2023pcpnet} + Cylinder3D \cite{zhu2021cylindrical}} \\
        \end{tabular}
    }
    \label{tab:i_gmo_f_gmo}
    \vspace{-4pt}
\end{table}

\begin{table}[t]
    \centering
    \renewcommand\arraystretch{1}
    \definecolor{mycolor}{rgb}{0.64, 0.87, 0.93}
    \caption{Comparisons of \textbf{Fine-Grained GMO and GSO Forecasting} on nuScenes-Occupancy dataset.}
    \renewcommand\tabcolsep{3pt}
    \scalebox{0.93}{
        \begin{tabular}{l|ccc|ccc|c}
        \toprule
         & \multicolumn{3}{c|}{$\text{IoU}_{c}$} & \multicolumn{3}{c|}{$\text{IoU}_{f}$ (2\ s)} & $\tilde{\text{IoU}}_{f}$\\
         \cline{2-8}
         \multirow{-2}{*}{Method} & \rotatebox{60}{GMO} & \rotatebox{60}{GSO} & \rotatebox{60}{mean} & \rotatebox{60}{GMO} & \rotatebox{60}{GSO} & \rotatebox{60}{mean} & \rotatebox{60}{GMO}   \\
        \midrule
        OpenOccupancy-C \cite{wang2023openoccupancy} & 9.62 & 17.21 & 13.42 & 7.41 & 17.30 & 12.36 & 7.86\\
        $\rm{SPC}^\dagger$ & 5.85 & 3.29 & 4.57 & 1.08 & 1.40 & 1.24 & 1.12\\
        PowerBEV-3D \cite{li2023powerbev} & 5.91 & - & - & 5.25 & - & - & 5.49 \\
        OCFNet (Cam4DOcc) \cite{ma2024cam4docc} & 11.02 & 17.79 & 14.41 & 9.20 & 17.83 & 13.52 & 9.66 \\
        OccProphet \cite{chen2025occprophet} & 13.71 & 24.42 & 19.06 & 9.34 & 24.56 & 16.95 & 10.33 \\
        Drive-OccWorld \cite{yang2025driving} & 16.90 & 20.20 & 18.50 & 14.30 & 21.20 & 17.80 & 14.90 \\
        \hline
        \rowcolor{mycolor!25}
        {\method} (ours) & \textbf{17.50} & \textbf{23.10} & \textbf{20.30} & \textbf{15.20} & \textbf{24.50} & \textbf{19.90} & \textbf{15.70}\\
        \bottomrule
        \multicolumn{8}{l}{$\rm{SPC}^\dagger$: SurroundDepth \cite{wei2023surrounddepth} + PCPNet \cite{luo2023pcpnet} + Cylinder3D \cite{zhu2021cylindrical}} 
        \end{tabular}
    }
    \label{tab:f_gmo_gso}
    \vspace{-10pt}
\end{table}

\section{Experiments} \label{sec:exp}
\subsection{Experimental Setups}
\subsubsection{Datasets} We evaluate our method on the nuScenes \cite{caesar2020nuscenes} and nuScenes-Occupancy \cite{wang2023openoccupancy} datasets \label{dataset}, targeting both occupancy forecasting and planning. Among the 850 annotated scenes, 700 are allocated for training and the remainder for evaluation. Following the protocol in \cite{ma2024cam4docc, yang2025driving}, the model takes as input the images from two past frames and the current frame to predict future states across four timestamps for 4D occupancy forecasting. The occupancy labels cover a spatial range of $[-51.2 m, 51.2 m]$ along the x and y axes and $[-5 m, 3 m]$ along the z axis, with a voxel resolution of 0.2 m, yielding a grid size of $(512, 512, 40)$.

\subsubsection{Tasks and Metrics.}
We evaluate {\method} on vison-centric 4D occupancy forecasting tasks and planning task:
\textbf{(1) Inflated Occupancy Forecasting} \cite{ma2024cam4docc}: predicting future states of general movable objects (\textbf{GMO}) represented by dilated occupancy grids derived from bounding-box annotations in nuScenes.
\textbf{(2) Fine-grained Occupancy Forecasting}: using voxel-level annotations from nuScenes-Occupancy, covering both general movable objects (\textbf{GMO}) and general static objects (\textbf{GSO}). We also provide evaluation results on multiple movable objects (\textbf{MMO}) and multiple static objects (\textbf{MSO}).
\textbf{(3) End-to-end Planning}: open-loop trajectory planning on nuScenes.

For occupancy forecasting, we adopt mean IoU (mIoU), reporting $\text{mIoU}_c$ for the current frame $(t=0)$, $\text{mIoU}_f$ for future timestamps $(t \in [1,f])$, and a time-weighted $\tilde{\text{mIoU}}_f$ that emphasizes short-term prediction accuracy. Planning performance is measured by the L2 distance between predicted and GT trajectories as well as object collision rate.

\subsubsection{Implementation Details}
We take the BEVFormer-based encoder \cite {li2022bevformer} as the scene encoder $\mathcal{E}$. The resolution of BEV features is set to $200\times200$. The future predictor $\mathcal{W}$ consists of three transformer layers with a hidden dimension of 256. Following previous work \cite{hu2022st, yang2025driving}, we set $\lambda_{plan}$ and $\lambda_{coll}$ to 1.
We adopt an AdamW optimizer with an initial learning rate of 2e-4 and a cosine annealing scheduler to train our model on 8 NVIDIA A100 GPUs.

\begin{table*}[htbp]
\centering
\caption{Comparisons of \textbf{Fine-Grained MMO and MSO Forecasting} on nuScenes-Occupancy dataset. Per-class $\tilde{\text{IoU}}_{f}$ is also provided.}
\definecolor{mycolor}{rgb}{0.64, 0.87, 0.93}
\renewcommand\arraystretch{1}
\setlength{\tabcolsep}{2.7pt}
\scalebox{0.9}{
\begin{tabular}{l|cc|cccccccccccccccc>{\columncolor{gray!10}}c}
\toprule
Method & $\text{mIoU}_{c}$ & 
\makecell{$\text{mIoU}_{f}$ \\ (2.5s)} &
\rotatebox{60}{Barrier} & 
\rotatebox{60}{Bicycle} & 
\rotatebox{60}{Bus} & 
\rotatebox{60}{Car} & 
\rotatebox{60}{Construction} & 
\rotatebox{60}{Motorcycle} & 
\rotatebox{60}{Pedestrian} & 
\rotatebox{60}{Traffic cone} & 
\rotatebox{60}{Trailer} & 
\rotatebox{60}{Truck} & 
\rotatebox{60}{Drivable surface} & 
\rotatebox{60}{Other} & 
\rotatebox{60}{Sidewalk} & 
\rotatebox{60}{Terrain} & 
\rotatebox{60}{Manmade} & 
\rotatebox{60}{Vegetation} & $\tilde{\text{mIoU}}_{f}$ \\
\toprule
Drive-OccWorld~\cite{yang2025driving}
& 12.9 & 11.5 & 11.5 & 6.9 & 10.2 & 13.2 & 7.7 & 8.3 & 7.6 
& 6.5 & 4.8 & 10.5 & 29.1 & 20.1 & 19.2 & 16.1 & 6.4 & 10.8 & 11.8 \\
\rowcolor{mycolor!25}{\method} (ours) 
& 15.2 & 14.4 & 14.8 & 9.2 & 12.7 & 15.8 & 9.4 & 9.7 & 9.0 
& 9.7 & 8.1 & 13.2 & 32.5 & 23.8 & 21.5 & 19.4 & 10.3 & 14.4 & 14.6 \\
\bottomrule
\end{tabular}
}
\label{tab:mmo_mso}
\end{table*}

\begin{table*}[htbp]
\setlength{\tabcolsep}{6pt}
\center
\caption{\textbf{End-to-end Planning Performance} on nuScenes validation. The ego status was not utilized in the planning module.}
\renewcommand\arraystretch{1}
\definecolor{mycolor}{rgb}{0.64, 0.87, 0.93}
\scalebox{0.95}{
\begin{tabular}{l|c|c|ccc>{\columncolor{gray!10}}c|ccc>{\columncolor{gray!10}}c}
\toprule
\multirow{2}{*}{Method} & \multirow{2}{*}{Input} & \multirow{2}{*}{ Aux.Sup.} & \multicolumn{4}{c|}{L2 (m) $\downarrow$} & \multicolumn{4}{c}{Collision (\%) $\downarrow$} \\
& & & {1s} & {2s} & {3s} & {Avg.} & {1s} & {2s} & {3s} & {Avg.} \\
\midrule
IL~\cite{ratliff2006maximum} & LiDAR & None & 0.44 & 1.15 & 2.47 & 1.35 & 0.08 & 0.27 & 1.95 & 0.77 \\
NMP~\cite{zeng2019end} & LiDAR & Box \& Motion & 0.53 & 1.25 & 2.67 & 1.48 & 0.04 & 0.12 & 0.87 & 0.34 \\
FF~\cite{hu2021safe} & LiDAR & Freespace & 0.55 & 1.20 & 2.54 & 1.43 & 0.06 & 0.17 & 1.07 & 0.43 \\
EO~\cite{khurana2022differentiable} & LiDAR & Freespace & 0.67 & 1.36 & 2.78 & 1.60 & 0.04 & 0.09 & 0.88 & 0.33 \\
\midrule
BevFormer \cite{li2022bevformer}+OccWorld \cite{zheng2023occworld} & Camera & 3D-Occ & 0.43 & 0.87 & 1.31 & 0.87 & - & - & - & - \\
BevFormer~\cite{li2022bevformer}+Occ-LLM~\cite{xu2025occ} & Camera & 3D-occ & 0.26 & 0.67 & 0.98 & 0.64 & - & - & - & -\\
\midrule
ST-P3~\cite{hu2022st} & Camera & Map \& Box \& Depth & 1.33 & 2.11 & 2.90 & 2.11 & 0.23 & 0.62 & 1.27 & 0.71  \\
UniAD~\cite{hu2023planning} & Camera & Map \& Box \& Motion \& Tracklets \& Occ & 0.48 & 0.96 & 1.65 & 1.03 & {0.05} & {0.17} & 0.71 & 0.31 \\
VAD-Base~\cite{jiang2023vad} & Camera & Map \& Box \& Motion & {0.54} & {1.15} & {1.98} & {1.22} & 0.04 & 0.39 & 1.17 & 0.53 \\
OccNet~\cite{tong2023scene} & Camera & 3D-Occ \& Map \& Box & 1.29 &2.13 &2.99 &{2.14} & 0.21 & 0.59 &1.37 &{0.72} \\
GenAD \cite{zheng2024genad} & Camera & Map \& Box \& Motion & 0.36 & 0.83 & 1.55 & 0.91 & 0.06 & 0.23 & 1.00 & 0.43 \\
UAD~\cite{guo2024end} & Camera & Box & 0.39 & 0.81 & 1.50 & 0.90 & 0.01 & \textbf{0.12} & 0.43 & 0.19 \\
RenderWorld \cite{yan2025renderworld} & Camera & None & 0.48 & 1.30 & 2.67 & 1.48 & 0.14 & 0.55 & 2.23 & 0.97 \\
SSR~\cite{li2024navigation} & Camera & None &  0.24 & 0.65 & 1.36 & 0.75 & \textbf{0.00} & 0.10 & 0.36 & \textbf{0.15} \\

PARA-Drive~\cite{weng2024drive} & Camera & Map \& Box \& Motion \& Tracklets \& Occ & 0.40 & 0.77 & 1.31 & 0.83 & 0.07 & 0.25 & 0.60 & 0.30 \\
GaussianAD \cite{zheng2024gaussianad} & Camera & 3D-Occ \& Map \& Box \& Motion & 0.40 & 0.64 & 0.88 & 0.64 & 0.09 & 0.38 & 0.81 & 0.42 \\
Drive-OccWorld \cite{yang2025driving} & Camera & 4D-Occ & 0.32 & 0.75 & 1.49 & 0.85 & 0.05 & 0.17 & 0.64 & 0.29 \\
\rowcolor{mycolor!25} {\method} (ours) & Camera & 4D-Occ & \textbf{0.23} & \textbf{0.51} & \textbf{0.85} & \textbf{0.53} & 0.14 & 0.20 & \textbf{0.16} & 0.17 \\
\bottomrule
\end{tabular}}
\label{tab:end2end}
\vspace{-8pt}
\end{table*}

\subsection{Main Results}
\subsubsection{Movable Objects 4D Occupancy Forecasting}
Table~\ref{tab:i_gmo_f_gmo} presents results on inflated and fine-grained GMO forecasting, focusing on dynamic objects prediction. Since dynamic agents are critical for driving safety, these results directly assess models' abilities to anticipate future motion patterns.

\textbf{Inflated GMO Forecasting}. As shown in the left part of Table~\ref{tab:i_gmo_f_gmo}, {\method} achieves the best performance with $\text{IoU}_c=40.8$, $\text{IoU}_f=37.2$, and $\tilde{\text{IoU}}_f=38.2$, surpassing the previous state-of-the-art Drive-OccWorld by about +1 point on all metrics. Compared with occupancy-only baselines like OpenOccupancy-C, our method delivers clear improvements, highlighting the benefit of richer spatiotemporal modeling. These results demonstrate that {\method} effectively captures both the current distribution and future evolution of moving objects, which is crucial for downstream planning.

\textbf{Fine-Grained GMO Forecasting}. As shown in the right part of Table~\ref{tab:i_gmo_f_gmo}, {\method} surpasses previous methods at the voxel level, achieving $\text{IoU}_c=16.2$ and $\text{IoU}_f=14.5$, with notable gains over OccProphet (+0.8 and +3.8). Although voxel-level prediction is more challenging, these improvements demonstrate the robustness of our design. {\method} captures not only coarse trajectories but also fine spatial details of dynamic agents, providing more reliable scene representations for safety-critical driving.

\subsubsection{Movable \& Static Objects 4D Occupancy Forecasting}
Table~\ref{tab:f_gmo_gso} and~\ref{tab:mmo_mso} present results on fine-grained forecasting of both movable and static objects, a more comprehensive setting that requires jointly modeling dynamic agents and static structures for reliable scene understanding.

\textbf{Fine-Grained GMO \& GSO Forecasting}. As shown in Table~\ref{tab:f_gmo_gso}, {\method} achieves the best performance across all metrics. It attains $\text{IoU}_c=20.3$ at the current frame, surpassing Drive-OccWorld by 1.8 points. For future predictions, it further improves to $\text{IoU}_f=15.2$ (GMO) and $24.5$ (GSO), outperforming OccProphet and Drive-OccWorld by clear margins. The weighted future score $\tilde{\text{IoU}_f}$ also sets a new state-of-the-art, showing that {\method} not only forecasts dynamic objects accurately but also captures fine-grained layouts of static structures.

\textbf{Fine-Grained MMO \& MSO Forecasting}. As shown in Table~\ref{tab:mmo_mso}, {\method} outperforms Drive-OccWorld across all metrics in forecasting fine-grained movable and static objects under the multi-class setting. It achieves $\text{mIoU}_c=15.2$ and $\text{mIoU}_f=14.4$, surpassing the baseline by +2.3 and +2.9, while the weighted score $\tilde{\text{mIoU}}_f=14.6$ sets a new state-of-the-art. Per-class analysis confirms broad gains across movable (e.g., Car, Truck, Bus) and static classes (e.g., Drivable surface, Sidewalk, Vegetation), with notable improvements on small and dynamic categories such as Pedestrian (+1.4) and Traffic cone (+3.2), demonstrating {\method} captures both heterogeneous agent motion and fine-grained static context.

\subsubsection{End-to-End Planning}
Table~\ref{tab:end2end} presents results on open-loop end-to-end planning on the nuScenes validation. 
The results demonstrate that {\method} achieves the best overall results, reaching an average L2 error of 0.53 m, substantially lower than Drive-OccWorld (0.85 m) and other camera-based baselines such as PARA-Drive (0.83 m) and GaussianAD (0.64 m). In particular, {\method} reduces long-horizon error at 3s to 0.85 m, demonstrating stronger temporal consistency and trajectory stability. For safety, our method obtains an average collision rate of 0.17\%, markedly better than prior occupancy-based planners like Drive-OccWorld (0.29\%). These results highlight that {\method} generates more informative implicit representations for downstream decision making, yielding both accurate and safe trajectories.

\begin{table}[t] 
\setlength{\tabcolsep}{2pt}
\centering
\caption{Ablations on \textbf{combinations of forecasting and planning}. We provide \textbf{Fine-Grained MMO and MSO Forecasting} results on nuScenes-Occupancy validation.}
\renewcommand\arraystretch{1}
\definecolor{mycolor}{rgb}{0.64, 0.87, 0.93}
\scalebox{0.95}{
\begin{tabular}{c|ccc|cccc|c}
\toprule
\multirow{2}{*}{Variants} & \multirow{2}{*}{$\text{mIoU}_c$} & \multirow{2}{*}{\makecell{$\text{mIoU}_f$ \\ (2.5\,s)}} & \multirow{2}{*}{$\tilde{\text{mIoU}}_f$} & \multicolumn{4}{c|}{L2 (m) $\downarrow$} & \multirow{2}{*}{\makecell{Latency\\(ms)}} \\
& & & & {1s} & {2s} & {3s} & {Avg.}  \\ \cmidrule{1-9}
Tightly Coupled & 15.1 & 14.3 & 14.6 & 0.23 & 0.50 & 0.84 & 0.52 & 764\\
\rowcolor{mycolor!25} Semi-Coupled & 15.2 & 14.4 & 14.6 & 0.23 & 0.51 & 0.85 & 0.53 & 706 \\
Decoupled & 14.9 & 14.3 & 14.5 & 0.29 & 0.78 & 1.55 & 0.87 & 558\\
\bottomrule
\end{tabular}}
\label{ab:planner}
\vspace{-12pt}
\end{table}


\subsection{Visualizations}
In Fig. \ref{fig:visualization}, we present qualitative results for fine-grained occupancy forecasting and planning across consecutive frames. The visualizations show that {\method} produces reliable predictions under varying time-of-day conditions. It also accurately captures the motion trends of dynamic objects. In particular, predictions for static objects remain highly stable over time, highlighting the effectiveness of our implicit residual world models.

\subsection{Analysis}
We conduct ablation studies to analyze the influence of the key components.

\noindent\textbf{Interaction between Forecasting and Planning.} We present the 4D occupancy forecasting and planning results of different variants in Table~\ref{ab:planner}. The forecasting performance of the three variants is largely comparable, while their planning accuracy and efficiency differ significantly. We observe that (1) using occupancy to filter candidate trajectories appears unnecessary, as it has only a marginal effect on planning performance (tightly coupled vs. semi-coupled) while bringing extra latency; (2) future BEV features generated by world models substantially improves planning accuracy (Semi-Coupled vs. Decoupled), although it inevitably increases inference latency, which highlights the effectiveness of world models in improving planning accuracy; (3) world models also provide strong supervisory signals for trajectory learning. Even in the fully decoupled setting, it still achieves performance comparable to other methods in Table~\ref{tab:end2end}.

\begin{table}[t]
    \centering
    \renewcommand\arraystretch{1}
    \definecolor{mycolor}{rgb}{0.64, 0.87, 0.93}
    \caption{Ablations on the \textbf{temporal self-supervision (TSS)} on BEV features. We provide \textbf{Fine-Grained GMO and GSO Forecasting} results on nuScenes-Occupancy validation.}
    \renewcommand\tabcolsep{5pt}
    \scalebox{0.95}{
        \begin{tabular}{l|ccc|ccc|c}
        \toprule
         & \multicolumn{3}{c|}{$\text{IoU}_{c}$} & \multicolumn{3}{c|}{$\text{IoU}_{f}$ (2\ s)} & $\tilde{\text{IoU}}_{f}$\\
         \cline{2-8}
         \multirow{-2}{*}{Method} & \rotatebox{60}{GMO} & \rotatebox{60}{GSO} & \rotatebox{60}{mean} & \rotatebox{60}{GMO} & \rotatebox{60}{GSO} & \rotatebox{60}{mean} & \rotatebox{60}{GMO}   \\
        \midrule
        \rowcolor{mycolor!25}{\method} & 17.50 & {23.10} & {20.30} & {15.20} & {24.50} & {19.90} & {15.70}\\
        W/ TSS & 17.50 & {23.20} & {20.30} & {15.10} & {24.50} & {19.80} & {15.60}\\
        \bottomrule
        \end{tabular}
    }
    \label{ab:loss_tss}
\end{table}

\begin{table}[t] 
\setlength{\tabcolsep}{10pt}
\center
\caption{Ablations on the \textbf{action conditioning}. We provide \textbf{Fine-Grained MMO and MSO Forecasting} results on nuScenes-Occupancy validation.}
\renewcommand\arraystretch{1}
\definecolor{mycolor}{rgb}{0.64, 0.87, 0.93}
\begin{tabular}{cc|ccc}
\toprule
CA & Additon & $\text{mIoU}_c$ & {$\text{mIoU}_f$ (2\,s)} & {$\tilde{\text{mIoU}}_f$} \\ \cmidrule{1-5}
\cmark &  & 14.2 & 13.8 & 13.9 \\
\rowcolor{mycolor!25}  & \cmark & 14.3 & 13.9 & 14.1 \\
\bottomrule
\end{tabular}
\label{ab:ca}
\vspace{-10pt}
\end{table}

\begin{table}[t] 
\setlength{\tabcolsep}{4pt}
\center
\caption{Ablations on the \textbf{feature alignment (FA)}. We provide per-class ${\text{IoU}}_{f}$ results on nuScenes-Occupancy validation.}
\renewcommand\arraystretch{1}
\definecolor{mycolor}{rgb}{0.64, 0.87, 0.93}
\begin{tabular}{c|cccc}
\toprule
Variants & barrier & construction  & sidewalk  & driveable surface  \\ \cmidrule{1-5}
w/o FA & 14.0 & 8.5 & 20.7 & 32.0  \\
\rowcolor{mycolor!25} w/ FA & 14.4 (\textbf{+0.4})  & 8.9 (\textbf{+0.4})  & 21.1 (\textbf{+0.4})  & 32.2 (\textbf{+0.2})  \\
\bottomrule
\end{tabular}
\label{ab:fa}
\end{table}

\begin{table}[ht] 
\setlength{\tabcolsep}{8pt}
\center
\caption{Ablations on the \textbf{BEV feature resolution}. We provide \textbf{Fine-grained GMO Forecasting} results on nuScenes-Occupancy validation.}
\renewcommand\arraystretch{1}
\definecolor{mycolor}{rgb}{0.64, 0.87, 0.93}
\begin{tabular}{c|c|ccc}
\toprule
 Method & Resolution & {$\text{IoU}_c$} & {$\text{IoU}_f$ (2\,s)} & {$\tilde{\text{IoU}}_f$} \\ \cmidrule{1-5}
 Drive-OccWorld & 200$\times$200 & 13.60 & 12.00 &12.40 \\
 \rowcolor{mycolor!25}{\method} (ours) & 200$\times$200 & 16.20 & 14.50 & 15.00  \\
 {\method} (ours) & 128$\times$128 & 11.80 & 10.80 & 11.10   \\
\bottomrule
\end{tabular}
\label{ab:reso}
\vspace{-7pt}
\end{table}

\noindent\textbf{Impact of Condition Interface and Feature Alignment.} We evaluate the effect of action conditioning and the alignment module. The detailed results are summarized in Table~\ref{ab:ca}. Using addition for conditioning achieves better performance compared to cross-attention while reducing the parameters. Moreover, as shown in Table~\ref{ab:fa}, removing the alignment module leads to a drop in $\text{IoU}_f$ for typical static objects, indicating that proper alignment of predicted BEV features can slightly alleviate error accumulation over time.

\noindent\textbf{Exploration on Temporal Self-Supervision.} We further evaluate temporal self-supervision (TSS) on BEV features. During training, the scene encoder $\mathcal{E}$ extracts BEV features $\mathbf{{S}}^{obs}_{1:f-1}$ from future multiview images $\mathbf{I}_{1:f-1}$, and an L2 loss is applied to supervise the predicted BEV features $\mathbf{S}^{bev}_{1:f-1}$. As shown in Table~\ref{ab:loss_tss}, TSS provides only a marginal regularization effect on occupancy prediction. This indicates that our implicit residual world model, guided by 4D occupancy, already enforces strong temporal consistency, leaving limited benefit for additional temporal constraints.


\noindent\textbf{Impact of Different BEV Feature Resolutions.} Finally, we compare different BEV feature resolutions. Results in Table~\ref{ab:reso} showcase that our method consistently outperforms the Drive-OccWorld baseline under the same resolution, demonstrating the effectiveness of our implicit residual world model. Moreover, higher spatial resolution (200$\times$200) provides a significant advantage over coarser features (128$\times$128), underlining the importance of fine-grained BEV representation for accurate occupancy prediction.

\section{Conclusion}
We presented {\method},  an end-to-end framework that shifts from reconstructing full future scenes to explicitly modeling their changes. {\method} builds a strong BEV representation of the current state from raw visual inputs, then uses the previous timestep’s BEV as a spatiotemporal prior to predict only action-conditioned residuals. We deliver that forecasting BEV features with occupancy as supervision markedly boost planning accuracy despite added latency, while using occupancy to filter candidate trajectories offers only marginal gains. At the same time, the world model provides strong supervisory signals for trajectory learning, enabling a fully decoupled variant to achieve performance comparable to recent methods with low latency.

\section{Acknowledgment}
This research was supported by Zhejiang Provincial Natural Science Foundation of China under Grant No.LQN25F030006.

\bibliographystyle{IEEEtran}
\bibliography{IEEEabrv,ref}

\end{document}